\documentclass[11pt]{article}

\usepackage[a4paper,margin=1in]{geometry}
\usepackage[utf8]{inputenc}
\usepackage[T1]{fontenc}
\usepackage[protrusion=true,expansion=false]{microtype}
\usepackage{amsmath,amssymb,amsfonts}
\usepackage{booktabs}
\usepackage{tabularx}
\usepackage{longtable}
\usepackage{array}
\usepackage{multirow}
\usepackage{makecell}
\usepackage{graphicx}
\usepackage{xcolor}
\usepackage[hidelinks,colorlinks=true,linkcolor=blue!50!black,citecolor=blue!50!black,urlcolor=blue!50!black]{hyperref}
\usepackage{url}
\usepackage{enumitem}
\usepackage{listings}
\usepackage{caption}
\usepackage{titlesec}
\usepackage{natbib}
\usepackage{authblk}
\usepackage{adjustbox}
\usepackage{pdflscape}
\usepackage{tikz}
\usepackage{pgfplots}
\pgfplotsset{compat=1.18}
\usepackage{algorithm}
\usepackage{algorithmic}
\usepackage{float}
\usepackage{subcaption}
\graphicspath{{images/}}

\titlespacing*{\section}{0pt}{1.5ex plus .2ex}{0.8ex plus .2ex}
\titlespacing*{\subsection}{0pt}{1.2ex plus .2ex}{0.6ex plus .2ex}

\title{\textbf{Opir: Efficient Multi-Task Safety Classification for Toxicity, Jailbreaks, Hate Speech, and Harmful Content}}

\author[1]{Ihor Stepanov\thanks{ingvarstep@knowledgator.com}}
\author[2]{Aleksandr Smechov\thanks{aleks@wordcab.com}}
\affil[1]{Knowledgator, Kyiv, Ukraine}
\affil[2]{Wordcab, New York, USA}
\date{}

\begin{document}

\maketitle

\begin{abstract}
\noindent
Real-time safety filtering for large language model (LLM) applications requires classifiers that can detect unsafe prompts, toxic language, jailbreak attempts, and unsafe responses without the cost profile of large guardrail models, and that can distinguish benign sensitive text from genuinely covert harmful content. In this paper, we introduce \textbf{Opir}, a family of encoder-based guardrail models built on the GLiClass architecture. Opir includes multi-task models for binary safe/unsafe classification, multi-label toxicity classification, jailbreak classification, and zero-shot unsafe prompt and response categorization. We also release edge variants with fewer than 100M parameters dedicated to binary safe/unsafe categorization. The models are trained on a three-level taxonomy containing 996 categories across 16 top-level labels, 126 mid-level labels, and 854 leaf labels. Opir's training data combines taxonomy-grounded unsafe prompts, adversarially mined hard negatives, benign safety-preserving examples, generated response examples, multilingual translations, and portions of the Aegis2 and WildGuard training subsets. We also open-sourced an evaluation harness that supports GLiClass and GLiNER2 backends as well as decoder-based models, and covers binary safety classification, multi-label categorization, toxicity, jailbreak detection, prompt safety, response safety, response refusal, and prompt subcategory views across public benchmark families. Across an expanded comparison spanning 12 safety-classification tasks and 17 category tasks against eight contemporary guardrail systems---including both GLiNER2-based and generative guardrail models---Opir variants are competitive on or ahead of the strongest open-weight baselines on the majority of benchmark datasets while operating with a substantially smaller deployment footprint. Latency measurements indicate that encoder variants can run with sub-30~ms p50 latency at 1024 tokens in the reported setup, while the smallest edge model achieves p50 latency below 10~ms.
\end{abstract}

\noindent\textbf{Keywords:} LLM safety; guardrails; toxicity classification; hate speech detection; jailbreak detection; prompt injection; harmful-content moderation; response safety; GLiClass; sequence classification; multi-label classification; safety taxonomy; multilingual moderation.

\section{Introduction}
\label{sec:intro}

Large language models~\citep{brown2020language,touvron2023llama,grattafiori2024llama3} are deployed in chat, agentic, and middleware settings where both user inputs and model outputs require real-time moderation. As LLMs gain more autonomy, the guardrail layer between an agent and a user becomes increasingly critical~\citep{shi2025lessons,greshake2023not}. Existing safeguard models such as Llama~Guard~\citep{inan2023llama}, WildGuard~\citep{han2024wildguard}, PolyGuard~\citep{kumar2025polyguard}, ShieldGemma~\citep{zeng2024shieldgemma}, Granite Guardian~\citep{padhi2024granite}, and the Llama-3.1-Nemotron Safety Guard family~\citep{ghosh2024aegis,ghosh2025aegis2} are typically large autoregressive models with 7B--22B parameters. While these models offer flexible, prompt-conditioned classification, their cost profile makes large-scale deployment expensive and adds significant per-request latency because each guarded interaction triggers one or more additional LLM forward passes. Many safety systems also emphasize binary safe/unsafe decisions. While these are practical for safety routing, real-world use cases often require finer categorization across toxicity, jailbreaks, prompt injection, and broader harmful-content domains~\citep{markov2023holistic,wang2024sorry,lin2023toxicchat}.

Opir tackles this with an encoder-based GLiClass~\citep{stepanov2025gliclass} model family. GLiClass extends the GLiNER~\citep{zaratiana2023gliner} architecture, originally proposed for zero-shot named entity recognition, to sequence classification by jointly encoding the input text and candidate labels with a bidirectional encoder. This enables zero-shot multi-label classification at a fraction of the cost of large generative guardrail models, and removes the brittleness of cross-encoder rerankers that must process each label-text pair sequentially. The Opir models are designed for multi-task moderation over prompt and response inputs, including binary safe/unsafe classification, toxicity classification, jailbreak classification, and zero-shot unsafe categorization over a hierarchical taxonomy. The project includes English-only, multilingual, and edge-oriented variants, allowing deployment across cloud-scale moderation services and single-machine edge devices.

\subsection{Contributions}
\label{sec:contributions}

We make the following contributions.

\begin{enumerate}[leftmargin=*,itemsep=2pt]
  \item \textbf{A three-level safety taxonomy} comprising 16 Level~1 categories, 126 Level~2 categories, and 854 Level~3 leaf labels (996 labels in total). The taxonomy covers ordinary toxicity, LLM-specific attacks (including instruction-hierarchy violations~\citep{wallace2024instruction} and indirect prompt injection~\citep{greshake2023not}), harmful-content categories, benign sensitive contexts, and uncertain boundary cases.
  \item \textbf{A GLiClass-based guardrail model family.} We develop Opir variants based on DeBERTaV3~\citep{he2021debertav3}, mDeBERTaV3, and compact encoders from the Ettin~\citep{ettin2025} and mmBERT~\citep{mmbert2025} families, with multi-task and edge-oriented deployment profiles.
  \item \textbf{A synthetic and open-data recipe.} We construct data from taxonomy-derived unsafe prompts, adversarial hard-negative mining inspired by red-teaming pipelines~\citep{perez2022red,zou2023universal,mehrotra2024tree}, benign safety-preserving contrast examples, generated responses, multilingual translation, and a number of open-source training datasets such as Aegis2~\citep{ghosh2025aegis2} and WildGuardMix~\citep{han2024wildguard}.
  \item \textbf{A multi-view evaluation harness.} We evaluate safety, toxicity, jailbreak detection, and main-category classification using GLiClass, GLiNER2, and vLLM~\citep{kwon2023efficient} backends, supporting prompt safety, response safety, response refusal, and prompt subcategory tasks across public benchmark families including HarmBench~\citep{mazeika2024harmbench}, JailbreakBench~\citep{chao2024jailbreakbench}, BeaverTails~\citep{ji2023beavertails}, PKU-SafeRLHF~\citep{ji2024pkusaferlhf}, XSTest~\citep{rottger2024xstest}, OR-Bench~\citep{cui2024orbench}, SimpleSafetyTests~\citep{vidgen2023simplesafetytests}, ToxicChat~\citep{lin2023toxicchat}, the OpenAI moderation benchmark~\citep{markov2023holistic}, WildGuardMix~\citep{han2024wildguard}, and PolyGuardPrompts~\citep{kumar2025polyguard}.
  \item \textbf{An extensive benchmark and latency comparison.} We compare Opir-multitask-large, Opir-multitask-multilang, Opir-edge, and Opir-edge-multilang with eight contemporary guardrail models across 12 safety datasets and 17 categorization datasets, and report first-pass throughput and p50/p95 latency measurements for GLiClass, GLiNER2, and vLLM guardrail backends. 

You can find evaluation code, scripts for reproducing the benchmark tables, and supplementary materials in the project repository.\footnote{
\url{https://github.com/Knowledgator/Opir}
}
\end{enumerate}

\section{Related Work}
\label{sec:related}

The literature on LLM safety classification spans more than a decade of work on toxicity detection, hate speech recognition, content moderation, jailbreaking, and adversarial robustness. In this section we situate Opir within five overlapping threads: classical content moderation, LLM-based guardrails, jailbreak and prompt-injection detection, multilingual safety, and efficient encoder-based classification.

\subsection{From Classical Toxicity Detection to LLM Moderation}

Early content moderation systems were built around supervised classifiers trained on social-media data, most prominently the Jigsaw \emph{Perspective API}~\citep{lees2022perspective} and the OpenAI moderation classifier~\citep{markov2023holistic}. \citet{markov2023holistic} introduced the \emph{holistic} approach: a unified taxonomy spanning hate, sexual content, self-harm, violence, harassment and related categories, learned with a multi-task transformer over both public and proprietary data. The OpenAI moderation benchmark released alongside that work remains a widely used evaluation set. Specialized BERT-style classifiers such as HateBERT~\citep{caselli2021hatebert} and ToxDectRoBERTa~\citep{zhou2021toxdect} were trained on platform-specific corpora and remain strong baselines for short-form toxic language. However, \citet{lin2023toxicchat} showed that toxicity classifiers trained on social-media data degrade substantially on real-world user--AI chat: their \textsc{ToxicChat} benchmark, drawn from the Vicuna online demo, exposes a large distribution shift between forum-style toxicity and the more conversational, role-play-heavy, instruction-following style of LLM users. ToxicChat additionally includes a \emph{jailbreak} label, anticipating the shift toward LLM-specific attack types.

A parallel thread has investigated dataset-level safety scaffolding for LLMs themselves. \citet{ji2023beavertails} released \textsc{BeaverTails}, a 330K-sample QA dataset annotated for harmlessness across 14 categories, with a focus on red-team-style prompts; \textsc{PKU-SafeRLHF}~\citep{ji2024pkusaferlhf} extends this with 44.6K refined prompts and 265K QA pairs labeled across 19 harm categories and three severity levels, supporting both response-safety classification and preference learning. \citet{rottger2024xstest} introduced \textsc{XSTest} to probe \emph{over-refusal}---safe prompts that look superficially unsafe---which exposes the brittle decision boundaries of many classifiers and aligned LLMs. \textsc{SimpleSafetyTests}~\citep{vidgen2023simplesafetytests} adds a small but pointed diagnostic set of high-priority harms, and \textsc{OR-Bench}~\citep{cui2024orbench} systematically constructs over-refusal probes at scale. All of these resources are included in our evaluation suite (Section~\ref{sec:eval}).

\subsection{LLM-Based Guardrails: Llama Guard, Aegis, WildGuard, PolyGuard, ShieldGemma, Granite Guardian, Qwen3Guard}

The current generation of safety classifiers is dominated by LLM-based guardrails that reuse general-purpose models and rely on prompt-conditioned classification.\textbf{Llama~Guard}~\citep{inan2023llama} pioneered this approach by fine-tuning Llama~2-7B on a curated safety dataset organized around a six-category taxonomy and emitting structured safe/unsafe verdicts conditioned on a policy prompt. Llama~Guard~2~\citep{llamaguard2024} and Llama~Guard~3 (released alongside Llama~3.1~\citep{grattafiori2024llama3}) expand the taxonomy and improve robustness to adversarial inputs.

NVIDIA's \textbf{Aegis}~\citep{ghosh2024aegis} fine-tunes Llama Guard on the proprietary \emph{Aegis Content Safety Dataset} with 13 risk categories and proposes an online no-regret ensemble of safety experts at inference time. \textbf{Aegis~2.0}~\citep{ghosh2025aegis2} extends this with the larger Nemotron Content Safety Dataset~v2 (formerly Aegis~2.0), a refined 12-category core taxonomy with nine fine-grained risks, and the Llama-3.1-Nemotron-Safety-Guard~8B models we benchmark against in Section~\ref{sec:eval}. NVIDIA has more recently released Nemotron-Content-Safety-Reasoning-4B~\citep{nemotronreasoning2025}, which supports both a low-latency classification mode and an optional reasoning-trace mode for custom policy enforcement.

The Allen Institute for AI's \textbf{WildGuard}~\citep{han2024wildguard} pursues a one-stop tool that simultaneously addresses prompt harmfulness, response harmfulness, and refusal detection. WildGuard's training data, \textsc{WildGuardMix}, contains 92K labeled examples covering 13 risk categories and explicitly mixes vanilla prompts, adversarial jailbreaks, and refusal/compliance responses; \citet{han2024wildguard} report that WildGuard closes the gap with GPT-4 on safety moderation in open-source settings. Building on WildGuardMix, \textbf{PolyGuard}~\citep{kumar2025polyguard} addresses the multilingual gap with \textsc{PolyGuardMix}, a 1.91M-sample training corpus spanning 17 languages, and \textsc{PolyGuardPrompts}, a 29K-sample evaluation benchmark; the corresponding PolyGuard-Qwen and PolyGuard-Qwen-Smol models are reported to outperform open-weight baselines by 5.5\% on average across multilingual safety benchmarks.

Google's \textbf{ShieldGemma}~\citep{zeng2024shieldgemma} and \textbf{ShieldGemma~2}~\citep{shieldgemma2_2025} extend the Gemma~\citep{gemmateam2024gemma} family with safety-specific fine-tuning and (in ShieldGemma~2) image moderation. IBM's \textbf{Granite Guardian}~\citep{padhi2024granite} integrates with the broader Granite~\citep{ibmgranite2024} stack and adds RAG-specific hallucination and grounding checks alongside conventional harm categories. Alibaba's \textbf{Qwen3Guard}~\citep{qwen3guard2025} provides a Qwen~3-based safety classifier that we include in our comparison, with both generative and classification variants. Finally, \textbf{AprielGuard}~\citep{aprielguard2025} explicitly targets input--output guardrail deployment with a hybrid training mixture of Salad-Bench~\citep{li2024salad}, in-the-wild jailbreak prompts~\citep{shen2023anything}, and WildGuardMix~\citep{han2024wildguard}.

All of these systems share the same architectural template: a decoder-only LLM is fine-tuned to emit structured verdicts, typically conditioned on a policy prompt. This template has well-known costs. First, inference latency is dominated by sequential token generation, which scales poorly to high-throughput moderation pipelines (e.g., agent loops, tool-use chains, RAG retrieval). Second, taxonomy changes require either prompt-engineering against the underlying LLM or re-fine-tuning, since the label set is encoded in natural language inside the policy prompt. Third, distillation and quantization can shrink the model but rarely below the 1B-parameter floor without significant accuracy loss. Encoder-based approaches such as \textsc{GLiGuard} ~\citep{zaratiana2026gliguardschemaconditionedclassificationllm} (a 300M-parameter GLiNER2-based safety classifier from Fastino) and \textsc{Gliner-Guard-Omni}~\cite{minko2026glinerguardunifiedencoder} attempt to address these costs but trade off coverage and accuracy. Opir occupies the same niche but extends the design with explicit multi-task heads, a richer 996-label taxonomy, and a multilingual variant.

\subsection{Jailbreaks, Prompt Injection, and Adversarial Robustness}

LLM-specific attacks decompose into two broad families: \emph{jailbreaks}, which manipulate the prompt to bypass alignment, and \emph{prompt injection}, which embeds adversarial instructions in third-party content~\citep{willison2022prompt,greshake2023not}. Universal adversarial suffixes were demonstrated by \citet{zou2023universal}; semantic and persona-based attacks were systematized by \citet{liu2024autodan} and \citet{chao2024pair}; and tree-of-attacks search was proposed by \citet{mehrotra2024tree}. \citet{shen2023anything} catalogued ``in-the-wild'' jailbreak prompts scraped from Reddit and Discord, which now feed many training sets.

Standardized evaluation has emerged around two benchmarks. \textbf{HarmBench}~\citep{mazeika2024harmbench} provides a unified red-teaming benchmark spanning 510 behaviors and 18 attack methods across copyright, cybercrime, CBRN, and other harm domains, along with a fine-tuned classifier. \textbf{JailbreakBench}~\citep{chao2024jailbreakbench} introduces an evolving repository of adversarial artifacts, a standardized threat model, and a 100-behavior dataset (JBB-Behaviors) aligned with OpenAI's usage policies; we use both the safety and behavior/category splits of JBB-Behaviors in Section~\ref{sec:eval}. \textbf{SALAD-Bench}~\citep{li2024salad} adds attack-enhanced prompts produced by human red-teamers, LLM-based red-teaming, and gradient attacks. The OWASP LLM Top~10~\citep{owaspllm2025} and Garak~\citep{derczynski2024garak} provide complementary industry-facing perspectives, the latter as an automated vulnerability scanner.

\emph{Indirect prompt injection}, first systematized by \citet{greshake2023not}, has emerged as a distinct threat for agentic LLMs. Recent classifiers~\citep{abdelnabi2024track} and benchmarks~\citep{debenedetti2024agentdojo} target the agent setting, where instructions may arrive via webpages, emails, calendar events, or repository files. The Opir taxonomy treats indirect prompt injection as a first-class Level~2 category under \texttt{ai\_system\_security\_and\_reliability}, with leaf labels for webpage, document, email, calendar, image, and repository injection vectors. Instruction-hierarchy attacks~\citep{wallace2024instruction} are similarly modeled as a top-level subcategory.

\subsection{Multilingual Safety}

English-centric guardrails fail to generalize to other languages, where translated harmful content can bypass safety filters and where culturally specific harms have no English analogue~\citep{deng2024multilingual,wang2024all}. \textbf{RTP-LX}~\citep{dewynter2024rtp} provides a translated toxicity benchmark across 28 languages; \textbf{PolyglotToxicityPrompts}~\citep{jain2024polyglot} extends RealToxicityPrompts~\citep{gehman2020realtoxicityprompts} to 17 languages. PolyGuard~\citep{kumar2025polyguard} is, to our knowledge, the strongest open multilingual guardrail at the time of writing. Opir's multilingual variants---Opir-multitask-multilang and Opir-edge-multilang---use mDeBERTaV3 and mmBERT backbones, respectively, and are trained on translations in 23 languages produced by DeepSeek-V3.1. We benchmark against PolyGuard on multilingual prompt and response safety in Section~\ref{sec:eval}.

\subsection{Efficient Encoder-Based Classification}

Beyond the safety domain, a growing line of work has tackled the latency cost of LLM classification by returning to encoder architectures. \textbf{SetFit}~\citep{tunstall2022efficient} fine-tunes sentence transformers~\citep{reimers2019sentence} with contrastive objectives and a logistic head, achieving strong few-shot performance. \textbf{GLiNER}~\citep{zaratiana2023gliner} introduced the now-canonical approach of jointly encoding text and candidate labels for zero-shot NER, eliminating the need to enumerate label-text pairs sequentially. \textbf{GLiNER Multi-task}~\citep{stepanov2024gliner_multitask} extends this to span-classification tasks beyond NER, including text classification, relation extraction, and question answering, and is the foundation for the GLiNER2 backend we evaluate against. \textbf{GLiClass}~\citep{stepanov2025gliclass} adapts the GLiNER architecture explicitly for sequence classification, reportedly running up to 50$\times$ faster than equivalent cross-encoders at comparable accuracy and supporting both zero-shot and few-shot scenarios. \textbf{GLiREL}~\citep{boylan2025glirel} applies the same template to relation extraction. Opir is built directly on top of GLiClass and inherits its uni-encoder, average-pooled, label-shuffled training recipe.

DeBERTaV3~\citep{he2021debertav3} remains a strong encoder backbone for moderation; it improves on DeBERTa~\citep{he2020deberta} with replaced-token-detection pre-training inspired by ELECTRA~\citep{clark2020electra}. The mDeBERTaV3 multilingual variant covers 100~languages. For edge deployments we use Ettin~\citep{ettin2025}, a 32M-parameter compact encoder family from JHU-CLSP, and mmBERT~\citep{mmbert2025}, its multilingual counterpart, which together provide the smallest practical encoder backbones for sub-10~ms inference at moderate sequence lengths.

\subsection{Positioning of Opir}

Opir sits at the intersection of these threads. Like Llama~Guard, Aegis, WildGuard, PolyGuard, and Qwen3Guard, it is a purpose-built safety classifier organized around a substantial taxonomy. Unlike them, it avoids decoder-only autoregression entirely, achieving order-of-magnitude latency reductions in our measurements. Like GLiGuard and Gliner-Guard-Omni, it uses an encoder-based GLiNER-family architecture; unlike them, it covers four task views (binary safety, toxicity, jailbreaks, zero-shot categorization), supports 23~languages, and is trained on a 996-label taxonomy that explicitly includes benign safety-preserving categories to suppress over-refusal on benchmarks such as XSTest and OR-Bench. The remainder of this paper documents the taxonomy (Section~\ref{sec:taxonomy}), model family (Section~\ref{sec:models}), data construction (Section~\ref{sec:data}), training (Section~\ref{sec:training}), augmentation (Section~\ref{sec:aug}), and evaluation results (Section~\ref{sec:eval}).

\section{Task Formulation}
\label{sec:task}

\begin{figure}[t]
\centering
\includegraphics[width=0.9\textwidth]{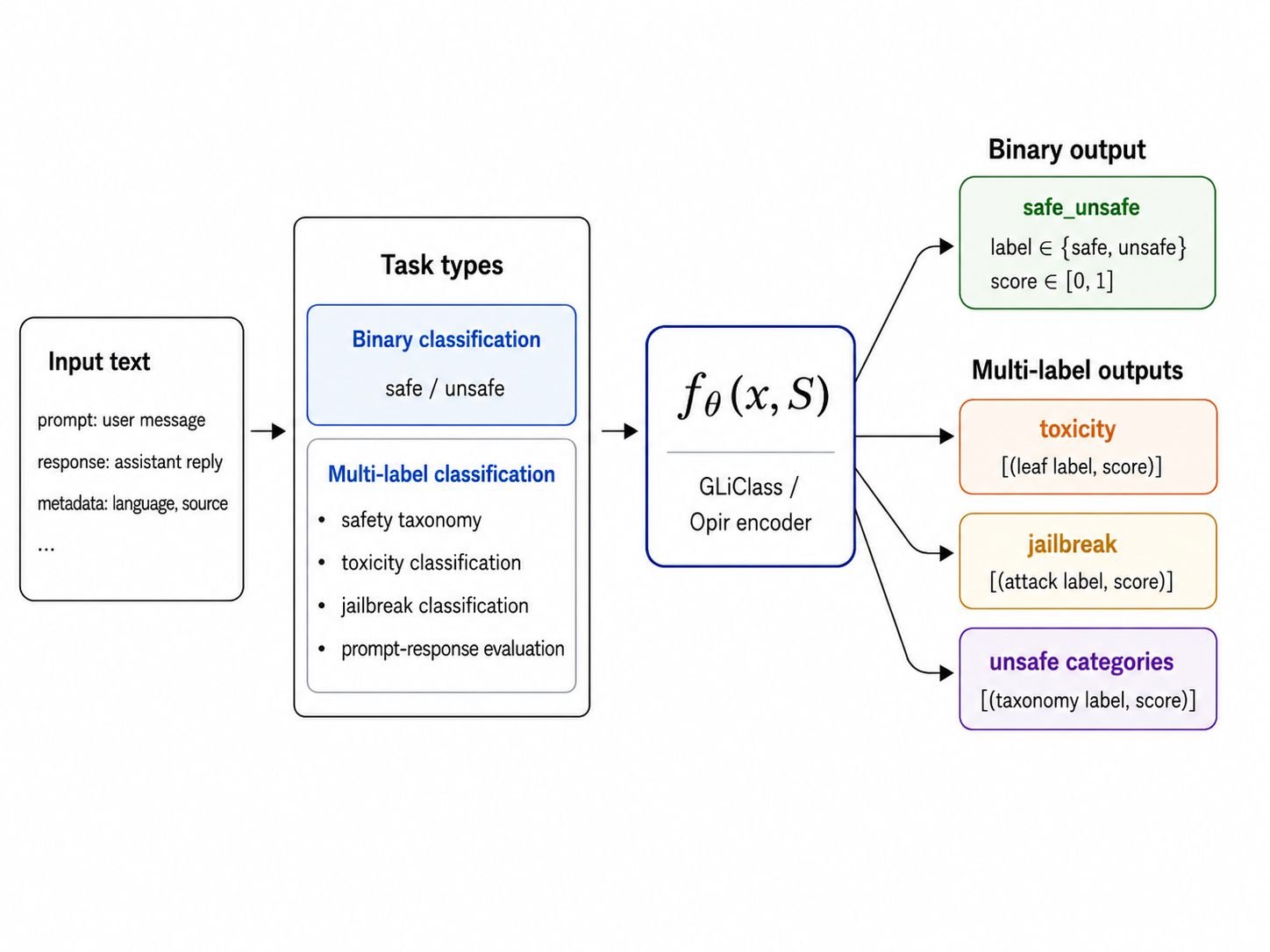}
\caption{Overview of Opir prediction tasks. Safe/unsafe classification is modeled as binary classification, while toxicity, jailbreak, and unsafe-category prediction are modeled as multi-label classification heads over task-specific label schemas.}
\label{fig}
\end{figure}

Given an input text $x$ and a list of labels$S$, Opir predicts safety labels for one or more subtasks. The input can be a user prompt, an assistant response, or a prompt--response pair, depending on the task settings. In our training configuration, inputs are encoded with a maximum sequence length of 4096 tokens. An overview of the task structure is shown in Figure~\ref{fig}.

Safe/unsafe classification is formulated as a binary classification task. Given an input prompt, response, or prompt--response pair, the model predicts whether the example is \emph{safe} or \emph{unsafe}, together with a confidence score. 

Toxicity classification is formulated as a multi-label task over conversational and social harms. The relevant taxonomy slice includes harassment and abuse, hate and discrimination, threats and intimidation, graphic or shocking content, abusive disruption, and psychological abuse or emotional harm. Because a single example may express multiple toxicity types, the model predicts a set of applicable toxicity labels rather than a single class.

Jailbreak classification is also modeled as multi-label prediction. This task captures LLM-specific adversarial behavior, including instruction-hierarchy attacks, secret or context exfiltration, tool and connector abuse, obfuscation and prompt smuggling, social-engineering attacks, indirect prompt injection, automation abuse, unsafe autonomy, tool-use risk, and related robustness or monitoring failures. The multi-label formulation allows a single attack to be assigned to multiple jailbreak patterns when appropriate.

Unsafe prompt and response categorization is modeled as multi-label classification over the broader safety taxonomy. This head supports routing, auditing, and fine-grained analysis beyond the binary safe/unsafe decision. The taxonomy covers top-level categories such as violence, self-harm, sexual content, child safety, privacy, cybersecurity, criminal activity, regulated goods and advice, biological, medical, and environmental harms, weapons of mass destruction, information manipulation, AI-system security, bias and fairness, uncertain cases, and safe or benign content.

\section{Safety Taxonomy}
\label{sec:taxonomy}

The Opir dataset is built around a three-level safety taxonomy. The top level contains 16 categories; the second level contains 126 categories; the third level contains 854 leaf labels; across all levels, the taxonomy contains 996 labels.

\begin{table}[ht]
\centering
\small
\caption{Top-level safety taxonomy (16 categories) with counts of Level~2 and Level~3 labels.}
\label{tab:taxonomy_summary}
\begin{tabular}{lrr}
\toprule
\textbf{Level~1 category} & \textbf{L2 cats.} & \textbf{L3 labels} \\
\midrule
toxicity & 6 & 41 \\
violence\_and\_physical\_harm & 5 & 30 \\
self\_harm\_and\_suicide & 5 & 30 \\
sexual\_content & 5 & 30 \\
child\_safety & 5 & 30 \\
personal\_information\_privacy\_and\_intellectual\_property & 18 & 129 \\
cybersecurity & 6 & 36 \\
criminal\_and\_illegal\_activity & 7 & 46 \\
regulated\_goods\_and\_advice & 6 & 33 \\
biological\_medical\_and\_environmental\_harm & 22 & 177 \\
weapons\_of\_mass\_destruction & 8 & 67 \\
information\_integrity\_and\_manipulation & 10 & 60 \\
ai\_system\_security\_and\_reliability & 12 & 79 \\
bias\_fairness\_and\_representation & 5 & 30 \\
other\_or\_uncertain & 2 & 12 \\
safe\_and\_benign & 4 & 24 \\
\midrule
\textbf{Total} & \textbf{126} & \textbf{854} \\
\bottomrule
\end{tabular}
\end{table}

The taxonomy includes both unsafe and safe/benign categories. This permits training examples that mention safety-sensitive concepts without requiring an unsafe label, such as counterspeech, harm prevention, defensive cybersecurity, general medical information, or appropriate refusal and redirection. Including explicit benign-sensitive categories is a known mitigation against over-refusal, which has been shown to be a primary failure mode of strict policy-prompted guardrails~\citep{rottger2024xstest,cui2024orbench}. The full Level~2 / Level~3 listing is reproduced in Appendix~\ref{app:taxonomy}.

\section{Model Family}
\label{sec:models}

Opir is a family of encoder-based GLiClass~\citep{stepanov2025gliclass} guardrail models. The documented variants are summarized in Table~\ref{tab:variants}.

\begin{table}[ht]
\centering
\small
\caption{Opir model variants.}
\label{tab:variants}
\begin{tabular}{lll}
\toprule
\textbf{Variant} & \textbf{Backbone} & \textbf{Role} \\
\midrule
Opir-multitask-large & DeBERTaV3-large~\citep{he2021debertav3} & multi-task safety classification \\
Opir-multitask-multilang & mDeBERTaV3-base~\citep{he2021debertav3} & multilingual multi-task \\
Opir-edge & Ettin-encoder-32m~\citep{ettin2025} & edge binary safe/unsafe \\
Opir-edge-multilang & mmBERT-small~\citep{mmbert2025} & multilingual edge binary safe/unsafe \\
\bottomrule
\end{tabular}
\end{table}

The multi-task variants are intended for safe/unsafe classification, toxicity classification, jailbreak classification, and zero-shot unsafe prompt/response categorization. The edge variants are intended for lower-cost binary safe/unsafe categorization, with the smallest model built on a 32M-parameter backbone. Initial checkpoints are seeded from publicly available GLiClass releases (\texttt{gliclass-instruct-large-v1.0}, \texttt{gliclass-x-base}, \texttt{gliclass-edge-v3.0}, and \texttt{gliclass-multilang-edge}) before safety-specific training.

\subsection{Architecture and Decoding}

Opir follows the GLiClass sequence-classification paradigm: the input text and a configurable set of candidate labels are jointly encoded by a bidirectional encoder. The GLiClass framework supports different approaches to pooling text and label representations, including average, first-token, last-token, max, and attention-style pooling. Likewise, label--text compatibility may be computed with dot-product similarity, bilinear scoring, cosine-style similarity, or a learned classification head.

\begin{figure}[H]
\centering
\includegraphics[width=0.92\textwidth]{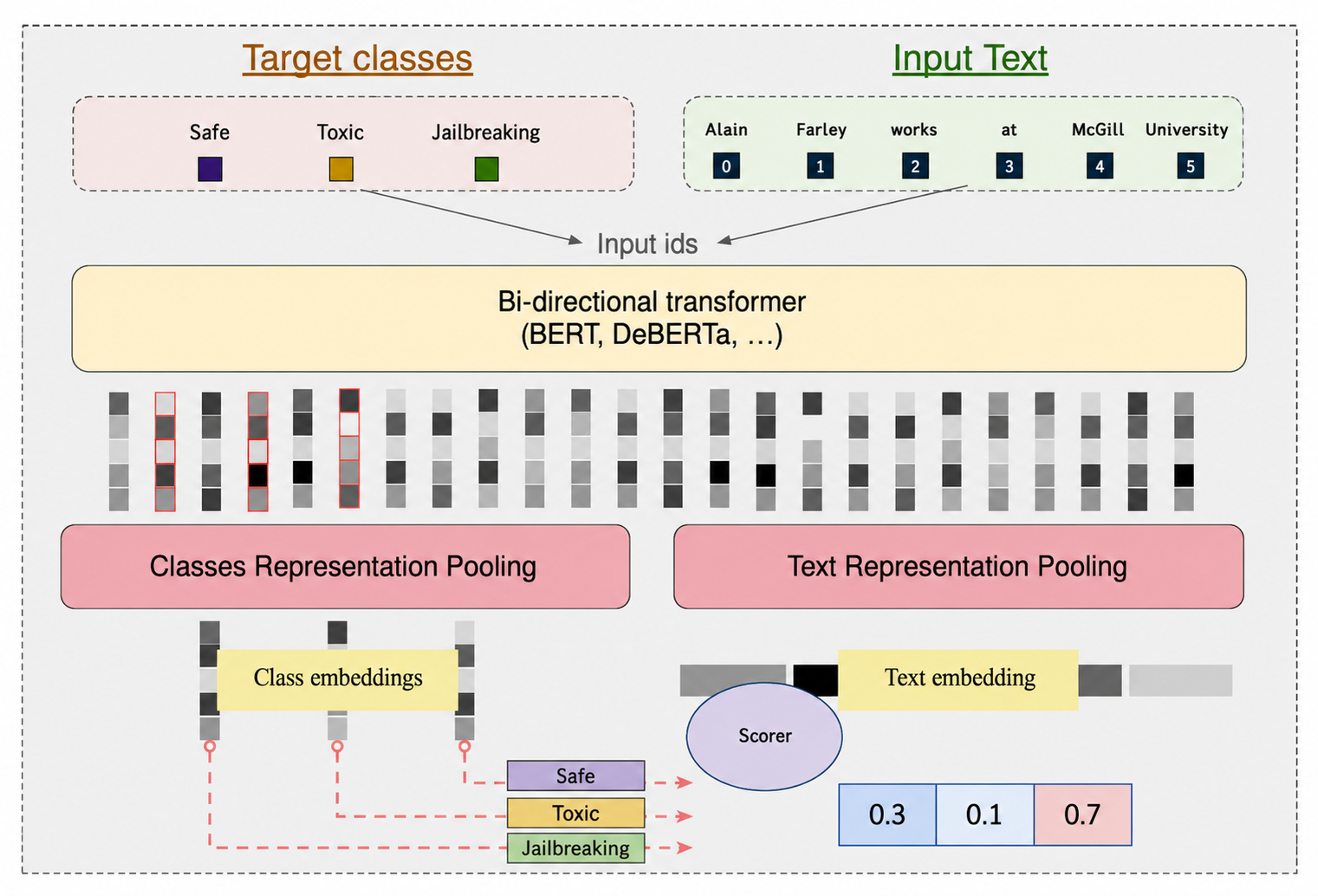}
\caption{Model architecture of Opir. Candidate labels and the input text are jointly encoded by a GLiClass-style bidirectional encoder. Task-specific pooling and scoring modules then produce logits for safe/unsafe classification, toxicity detection, jailbreak detection, and taxonomy-category prediction.}
\label{fig:model_architecture}
\end{figure}

Formally, given an input text $t$ and a candidate label set $L = \{\ell_1, \dots, \ell_k\}$, the encoder produces contextual representations
\begin{equation}
H = E_\theta(t, L).
\end{equation}
A text representation $z_t$ and label representations $z_{\ell_i}$ are obtained through configurable pooling functions:
\begin{equation}
z_t = P_{\text{text}}(H, t), \qquad
z_{\ell_i} = P_{\text{label}}(H, \ell_i).
\end{equation}
The model then computes one logit per candidate label,
\begin{equation}
a_i = g_\theta(z_t, z_{\ell_i}),
\end{equation}
where $g_\theta$ denotes the checkpoint-specific scorer or classification head.

Decoding depends on the task view. For multi-label tasks, such as toxicity, jailbreak, or taxonomy-category prediction, logits are converted to independent probabilities with a sigmoid function and labels are emitted when their scores exceed a configurable threshold $\tau$:
\begin{equation}
p_i = \sigma(a_i), \qquad \hat{y}_i = \mathbb{I}[p_i \geq \tau].
\end{equation}
For single-label tasks, such as binary safe/unsafe classification in the corresponding evaluation view, logits are normalized with a softmax and the highest-scoring class is selected:
\begin{equation}
p_i = \mathrm{softmax}(a)_i, \qquad \hat{y} = \arg\max_i p_i.
\end{equation}
Thus, sigmoid or softmax normalization is applied during post-processing according to whether the task is multi-label or single-label. Because the candidate label set is supplied at inference time, the same encoder can support fixed binary decisions as well as zero-shot classification over arbitrary safety taxonomies.

\section{Data Construction}
\label{sec:data}

For each node in the taxonomy, 30 unsafe prompts are generated by an LLM-as-author pipeline. Hard negatives are mined by evolutionarily modifying initial prompts to bypass existing safety models, in the spirit of Evol-Instruct~\citep{xu2024wizardlm} and recent automated red-teaming work~\citep{perez2022red,mehrotra2024tree}. LLMs are used as judges~\citep{zheng2023judging} to validate whether a prompt remains unsafe, using a panel of DeepSeek-V3.1, MiniMax-M2.5, and Meta-Llama-3.3-70B-Instruct models via the commercial SCX.ai API. Using a panel rather than a single judge follows the LLM-jury argument of \citet{verga2024replacing}, which reduces single-model bias in safety judgments.

\begin{figure}[H]
\centering
\includegraphics[width=0.92\textwidth]{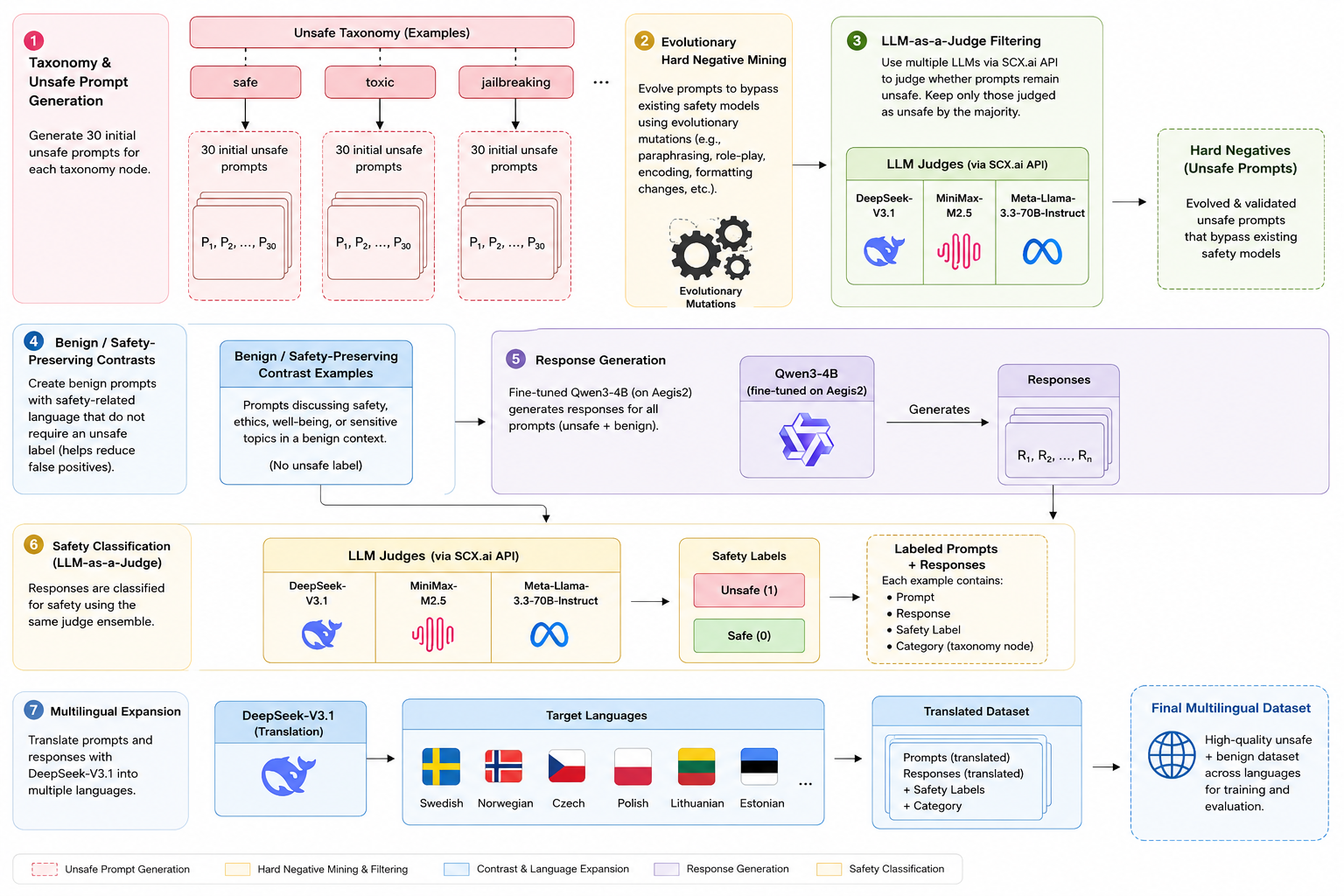}
\caption{Data construction pipeline for Opir. Taxonomy nodes seed unsafe prompt generation, hard-negative mining, benign-sensitive contrast construction, response generation and judging, multilingual translation, and final task-view formatting for training and evaluation.}
\label{fig:data_pipeline}
\end{figure}

Data construction also includes benign or safety-preserving contrast examples drawn from the taxonomy's \texttt{safe\_and\_benign} branch. These examples contain safety-related language but do not require an unsafe label, making them useful for reducing false positives on benign sensitive contexts---exactly the failure mode probed by XSTest~\citep{rottger2024xstest} and OR-Bench~\citep{cui2024orbench}.

To obtain response examples, a Qwen3-4B model~\citep{yang2024qwen3} is fine-tuned on Aegis2~\citep{ghosh2025aegis2} and used to generate responses for the generated prompts. Responses are then classified for safety with an LLM-as-a-judge pipeline. The multilingual dataset is produced by translating prompts and responses with DeepSeek-V3.1 into Swedish, Norwegian, Czech, Polish, Lithuanian, Estonian, Latvian, Spanish, Finnish, English, German, French, Romanian, Italian, Portuguese, Dutch, Ukrainian, Russian, Hindi, Chinese, Japanese, Korean, and Arabic (23 languages in total).

\begin{table}[ht]
\centering
\small
\caption{Dataset files used for training and post-training. \texttt{gliclass\_full\_*} files are used for training \texttt{Opir-multitask-large} and \texttt{Opir-multitask-multilang} model variants. \texttt{gliclass\_safety\_*} files are used for training the \texttt{Opir-edge} and \texttt{Opir-edge-multilang} model variants.}
\label{tab:datafiles}
\begin{tabular}{lrl}
\toprule
\textbf{File} & \textbf{Examples} & \textbf{Description} \\
\midrule
\texttt{gliclass\_safety\_multi.json} & 531{,}007 & multilingual safety examples \\
\texttt{gliclass\_safety\_en.json}    & 213{,}809 & English safety examples \\
\texttt{gliclass\_full\_multi.json}   & 1{,}106{,}635 & multilingual multi-task examples \\
\texttt{gliclass\_full\_en.json}      & 426{,}356 & English multi-task examples \\
\texttt{gliclass\_post\_training.json} & 18{,}000  & post-training examples \\
\bottomrule
\end{tabular}
\end{table}

The training data also includes portions of the Aegis2~\citep{ghosh2025aegis2} and WildGuardMix~\citep{han2024wildguard} training subsets. Aegis2 contributes 12-category labeled prompts curated from Anthropic HH-RLHF~\citep{bai2022training}; WildGuardMix contributes synthetic and in-the-wild jailbreak prompts paired with refusal/compliance responses. We follow the original licenses for both subsets.

\section{Training}
\label{sec:training}

Training is run using a Python script that loads the JSON training dataset, extends it with the \texttt{knowledgator/gliclass-v3-logic-dataset} to maintain general classification capability (a form of replay-based continual learning~\citep{lopez2017gradient}), shuffles the combined data, and uses a 90/10 train/evaluation split. Training is performed in two stages: initial training on the main dataset for 3 epochs, followed by post-training on a 10\% sample of examples after augmentation. Table~\ref{tab:hparams} lists the hyperparameters.

\begin{table}[ht]
\centering
\small
\caption{Training hyperparameters.}
\label{tab:hparams}
\begin{tabular}{ll}
\toprule
\textbf{Hyperparameter} & \textbf{Value} \\
\midrule
Problem type & multi\_label\_classification \\
Architecture type & uni-encoder \\
Pooling & average pooling \\
Class-token pooling & first token \\
Maximum sequence length & 1024 \\
Batch size & 8 \\
Gradient accumulation steps & 1 \\
Encoder learning rate & $1\times10^{-6}$ \\
Other/head learning rate & $3\times10^{-6}$ \\
Encoder weight decay & 0.01 \\
Other/head weight decay & 0.01 \\
Scheduler & cosine \\
Warmup ratio & 0.05 \\
Dropout & 0.3 \\
Label shuffling & enabled \\
Precision & bf16=True; fp16=False by default \\
Checkpoint save interval & 1000 steps \\
Checkpoint limit & 3 \\
Focal loss $\alpha$~\citep{lin2017focal} & 0.7 \\
Focal loss $\gamma$ & $-1$ \\
Focal loss reduction & none \\
Contrastive loss coefficient & 0.0 \\
\bottomrule
\end{tabular}
\end{table}

The training code also supports optional online Elastic Weight Consolidation~\citep{kirkpatrick2017overcoming} for continual learning, with $\lambda_{\mathrm{EWC}}=100.0$, $\gamma_{\mathrm{EWC}}=0.95$, and Fisher normalization enabled. This is intended for downstream fine-tuning scenarios where a deployer extends Opir with site-specific safety policies without forgetting the base taxonomy.

\section{Augmentation}
\label{sec:aug}

When augmentation is enabled, each training item can be modified probabilistically by removing labels, adding random labels from the label pool, prepending or appending non-overlapping example text, replacing labels with synonyms when metadata is available, inserting label descriptions, or inserting one or two few-shot examples with overlapping labels. These augmentations follow the recipe of label dropout and prompt perturbation common in zero-shot information-extraction training~\citep{stepanov2024gliner_multitask,bogdanov2024nuner}.

The post-training stage uses a 10\% sample of the main examples and applies these augmentations to improve robustness to label-set changes, prompt-formatting variation, and few-shot contexts. Additional prompt-injection augmentation samples rows from the dataset to insert safe-looking distractor instructions, such as label overrides and repeated safe tokens, and preserves metadata about the insertion offset and source. The motivation is similar to that of \citet{wallace2024instruction}: by exposing the classifier to instruction-hierarchy violations at training time, we hope to suppress instruction-following behavior that could be exploited by injected content.

\section{Evaluation}
\label{sec:eval}

\subsection{Evaluation Protocol}

The Opir evaluator script supports GLiClass, GLiNER2, and (via vLLM) decoder-based guardrails such as WildGuard~\citep{han2024wildguard}, the Llama-3.1-Nemotron-Safety-Guard family~\citep{ghosh2025aegis2}, PolyGuard~\citep{kumar2025polyguard}, and Qwen3Guard~\citep{qwen3guard2025}. All models run zero-shot classification with a configurable threshold, defaulting to 0.5.

For multi-label categorization, predictions and labels are binarized with the \texttt{MultiLabelBinarizer} class from the \texttt{scikit-learn} library; the evaluator reports micro, macro, and weighted F1. For binary safety datasets, predicted and gold labels are normalized to safe and unsafe. Evaluation reports accuracy, micro F1, macro F1, weighted F1, per-label precision/recall/F1/support, and predicted/gold label counts.

The evaluation suite spans the benchmark families referenced in Table~\ref{tab:benches}.

\begin{table}[ht]
\centering
\small
\caption{Benchmark families used in the evaluation suite.}
\label{tab:benches}
\begin{tabular}{ll}
\toprule
\textbf{Benchmark} & \textbf{Evaluation view} \\
\midrule
OpenAI moderation~\citep{markov2023holistic} & safety and category \\
Aegis / Aegis2~\citep{ghosh2024aegis,ghosh2025aegis2} & prompt safety, response safety, categories \\
SimpleSafetyTests~\citep{vidgen2023simplesafetytests} & safety \\
HarmBench~\citep{mazeika2024harmbench} & prompt and response safety \\
PKU-SafeRLHF~\citep{ji2024pkusaferlhf} & prompt/response safety \\
BeaverTails~\citep{ji2023beavertails} & safety \\
XSTest~\citep{rottger2024xstest} & safety / over-refusal \\
OR-Bench~\citep{cui2024orbench} & over-refusal (80k, hard-1k, toxic) \\
ToxicChat~\citep{lin2023toxicchat} & safety, toxicity, jailbreak \\
WildGuardMix~\citep{han2024wildguard} & prompt safety, response safety, refusal, subcategory \\
PolyGuardPrompts~\citep{kumar2025polyguard} & prompt safety, response safety, refusal, subcategory \\
JBB-Behaviors~\citep{chao2024jailbreakbench} & safety, behavior, category \\
PAN12 predator~\citep{pan12predator} & conversational safety \\
\bottomrule
\end{tabular}
\end{table}

Latency benchmarking reports throughput (samples/s) and p50/p95 request latency in milliseconds at sequence lengths 64, 256, 512, and 1024. The current latency log records model name, backend, sequence length, throughput, p50, and p95.

\subsection{Binary Safety Classification: Comparison Across 11 Guardrail Systems}
\label{sec:eval_safety}

Table~\ref{tab:safety_full} reports macro F1 on 12 binary safety datasets across 11 guardrail systems: GLiGuard-LLMGuardrails-300M (an encoder-based safety classifier from Fastino Labs), the four Opir variants (\texttt{Opir-multitask-large}, \texttt{Opir-multitask-multilang}, \texttt{Opir-edge}, \texttt{Opir-edge-multilang}), ~\citet{minko2026glinerguardunifiedencoder}'s Gliner-Guard-Omni, WildGuard~\citep{han2024wildguard} served via vLLM, Llama-3.1-Nemotron-Safety-Guard~v3 (denoted \emph{Nemotron Safety Guard v3})~\citep{ghosh2025aegis2}, PolyGuard-Qwen and PolyGuard-Qwen-Smol~\citep{kumar2025polyguard}, and Qwen3Guard-Gen-8B~\citep{qwen3guard2025}.

Across the 12 rows, \texttt{Opir-multitask-large} obtains the second-highest row-average macro F1 and wins two individual datasets (WildGuard prompt safety and JBB-Behaviors safety). \texttt{Opir-edge-multilang}---a multilingual binary classifier with under 100M parameters---wins Aegis prompt safety (0.9321) and WildGuard response safety (0.9194). Decoder-based guardrails remain strongest on several adversarial or benchmark-specific splits: Nemotron Safety Guard v3 leads on OAI safety, SafeRLHF response safety, ToxicChat safe/unsafe, and ToxicChat toxicity; PolyGuard-Qwen leads on both PolyGuard safety splits; Qwen3Guard-Gen-8B leads on Aegis response safety. The takeaway is that encoder-based Opir variants are competitive with 7B--8B decoder guardrails on average, while operating at a small fraction of their inference cost.

\begin{table}[H]
\centering
\scriptsize
\setlength{\tabcolsep}{2.4pt}
\renewcommand{\arraystretch}{0.98}
\caption[Macro F1 on binary safety classification datasets]{Macro F1 on 12 binary safety classification datasets across 11 guardrail systems. \textbf{Bold} indicates the best score per row; \underline{underline} indicates the second-best score. Higher is better.}
\label{tab:safety_full}
\begin{adjustbox}{width=\textwidth,center}
\begin{tabular}{l*{11}{r}}
\toprule
\textbf{Dataset} & \textbf{GG} & \textbf{O-ML} & \textbf{O-E} & \textbf{O-EM} & \textbf{O-MM} & \textbf{GGO} & \textbf{WG} & \textbf{NSG} & \textbf{PG-Q} & \textbf{PG-S} & \textbf{QG} \\
\midrule
oai\_safety                            & 0.6396 & 0.6075 & 0.5986 & 0.6397 & 0.6126 & 0.6785 & 0.7172 & \textbf{0.7676} & \underline{0.7277} & 0.6791 & 0.6706 \\
aegis\_prompt\_safety                  & 0.8161 & \underline{0.9308} & 0.8788 & \textbf{0.9321} & 0.8671 & 0.7225 & 0.7531 & 0.8433 & 0.8379 & 0.8278 & 0.8249 \\
aegis\_response\_safety                & 0.7648 & 0.7647 & 0.7916 & 0.8506 & 0.7739 & 0.7638 & 0.8377 & 0.7908 & \underline{0.8585} & 0.8423 & \textbf{0.8672} \\
saferlhf\_response\_safety             & 0.7476 & 0.8733 & 0.8261 & 0.8382 & 0.8327 & 0.7651 & \underline{0.9196} & \textbf{0.9243} & 0.8601 & 0.8293 & 0.7757 \\
wildguard\_prompt\_safety              & 0.8728 & \textbf{0.9791} & 0.8988 & \underline{0.9486} & 0.8884 & 0.7262 & 0.9037 & 0.8594 & 0.9029 & 0.8900 & 0.9095 \\
wildguard\_response\_safety            & 0.6413 & \underline{0.9164} & 0.8606 & \textbf{0.9194} & 0.8522 & 0.6438 & 0.8571 & 0.8695 & 0.8735 & 0.8526 & 0.6926 \\
polyguard\_prompt\_safety              & 0.8290 & 0.8116 & 0.5224 & 0.5873 & 0.6938 & 0.6926 & 0.8108 & 0.8432 & \textbf{0.9073} & 0.8740 & \underline{0.9069} \\
polyguard\_response\_safety            & 0.6079 & 0.8079 & 0.5516 & 0.6884 & 0.8150 & 0.6551 & 0.8032 & \underline{0.8668} & \textbf{0.8732} & 0.8249 & 0.7142 \\
toxicchat\_safe\_unsafe                & 0.5470 & 0.5730 & 0.5092 & 0.5489 & 0.5452 & 0.4899 & 0.5713 & \textbf{0.6323} & 0.5847 & \underline{0.5782} & 0.5701 \\
toxicchat\_toxicity                    & 0.7280 & \underline{0.8325} & 0.4260 & 0.6619 & 0.5370 & 0.7627 & 0.8129 & \textbf{0.8517} & 0.8237 & 0.8114 & 0.7977 \\
toxicchat\_jailbreaking                & 0.4357 & \underline{0.6634} & 0.0432 & 0.3951 & 0.1930 & \textbf{0.7054} & 0.5713 & 0.6323 & 0.5845 & 0.5786 & 0.5701 \\
jbb\_behaviors\_safety                 & 0.6672 & \textbf{0.8932} & 0.5783 & 0.7241 & 0.6072 & 0.4511 & 0.7583 & \underline{0.7917} & 0.6435 & 0.6460 & 0.6503 \\
\midrule
\textbf{Row average (12)}              & 0.6914 & \underline{0.8045} & 0.6238 & 0.7195 & 0.6857 & 0.6714 & 0.7647 & \textbf{0.8061} & 0.7898 & 0.7612 & 0.7458 \\
\textbf{Wins}                          & 0 & 2 & 0 & 2 & 0 & 1 & 0 & 4 & 2 & 0 & 1 \\
\bottomrule
\end{tabular}
\end{adjustbox}
\vspace{0.5ex}
\begin{minipage}{\textwidth}
\footnotesize
\textbf{Model abbreviations:} GG = GLiGuard-LLMGuardrails-300M; O-ML = Opir-multitask-large; O-E = Opir-edge; O-EM = Opir-edge-multilang; O-MM = Opir-multitask-multilang; GGO = Gliner-Guard-Omni; WG = WildGuard (vLLM); NSG = Nemotron Safety Guard v3; PG-Q = PolyGuard-Qwen; PG-S = PolyGuard-Qwen-Smol; QG = Qwen3Guard-Gen-8B.
\end{minipage}
\end{table}

Across these 12 rows, Nemotron Safety Guard v3 obtains the highest row-average macro F1 (0.8061), with \texttt{Opir-multitask-large} very close behind (0.8045) and PolyGuard-Qwen third (0.7898). \texttt{Opir-multitask-large} nevertheless wins two individual datasets and remains competitive with substantially larger decoder-based guardrails. Among the Opir variants, \texttt{Opir-multitask-large} provides the best accuracy, while \texttt{Opir-edge} and \texttt{Opir-edge-multilang} represent lower-latency binary classifiers for deployment-constrained settings.

\subsection{Safety Categorization Accuracy}
\label{sec:eval_cat}

Table~\ref{tab:category_full} reports per-row macro accuracy on the multi-label \emph{safety categorization} task across four encoder-based systems for which the full categorization output is available: GLiGuard-300M, \texttt{Opir-multitask-large}, \texttt{Opir-multitask-multilang}, and Gliner-Guard-Omni. Decoder-based guardrails (WildGuard, PolyGuard, Nemotron Safety Guard, Qwen3Guard) emit free-text rationales rather than full category vectors and are therefore excluded from this view; we would emphasize that this is a property of the model, not of Opir's evaluation harness.

\begin{table}[H]
\centering
\small
\caption{Categorization accuracy across 17 datasets / category splits. Higher is better. \textbf{Bold} indicates the best score per row.}
\label{tab:category_full}
\begin{tabular}{lrrrr}
\toprule
\textbf{Dataset / category split} & \textbf{GLiGuard} & \textbf{Opir-multitask} & \textbf{Opir-multitask} & \textbf{Gliner-} \\
                                  & \textbf{300M}     & \textbf{-large} & \textbf{-multilang} & \textbf{Guard-Omni} \\
\midrule
oai (OpenAI moderation)            & 0.4369 & \textbf{0.4767} & 0.3282 & 0.4390 \\
aegis\_categories                  & 0.2488 & \textbf{0.6284} & 0.5138 & 0.2289 \\
simplest                           & 0.7587 & \textbf{0.8668} & 0.8449 & 0.8138 \\
simplesafetytests                  & 0.7048 & \textbf{0.9138} & 0.8370 & 0.8606 \\
harmbench\_prompts                 & 0.1710 & \textbf{0.5432} & 0.4828 & 0.2986 \\
harmbench\_responses               & 0.2009 & \textbf{0.2726} & 0.2158 & 0.0294 \\
saferlhf                           & 0.3582 & \textbf{0.4835} & 0.3805 & 0.2756 \\
beavertails                        & 0.2230 & \textbf{0.4060} & 0.3196 & 0.3027 \\
xstest                             & 0.8335 & \textbf{0.9439} & 0.8149 & 0.6731 \\
pan12\_predator\_conv\_safety      & 0.3876 & \textbf{0.4736} & 0.4698 & 0.4481 \\
wildguard\_prompt\_subcategory     & 0.3909 & \textbf{0.8335} & 0.6717 & 0.3824 \\
polyguard\_prompt\_subcategory     & 0.3416 & 0.4796 & \textbf{0.5560} & 0.3159 \\
or\_bench\_80k                     & \textbf{0.7254} & 0.5032 & 0.4224 & 0.3202 \\
or\_bench\_hard\_1k                & \textbf{0.5353} & 0.3268 & 0.2660 & 0.0477 \\
or\_bench\_toxic                   & \textbf{0.5982} & 0.4058 & 0.4591 & 0.4973 \\
jbb\_behaviors\_behavior           & 0.0593 & 0.2576 & 0.7123 & \textbf{0.7217} \\
jbb\_behaviors\_category           & 0.2038 & 0.4178 & \textbf{0.5937} & 0.4693 \\
\midrule
\textbf{Row average (17)}          & 0.3987 & \textbf{0.5432} & 0.5230 & 0.4073 \\
\textbf{Wins}                      & 3      & 11              & 2      & 1 \\
\bottomrule
\end{tabular}
\end{table}

\texttt{Opir-multitask-large} wins 11 of 17 categorization rows and achieves the highest average accuracy (0.5432), with substantial margins on Aegis categories ($+0.38$ over GLiGuard-300M), HarmBench prompts ($+0.37$), WildGuard prompt subcategory ($+0.44$), and JBB-Behaviors category ($+0.21$ over GLiGuard-300M). \texttt{Opir-multitask-multilang}, the multilingual model variant, wins on the PolyGuard prompt subcategory (where multilingual coverage matters) and the JBB-Behaviors category. The OR-Bench family is the principal failure mode for the Opir-multitask models: here GLiGuard-300M's training mixture (which appears to include OR-Bench-style benign prompts directly) wins all three rows, while Opir's three-level taxonomy maps OR-Bench prompts onto its broader benign-sensitive categories with higher abstain rates, reducing accuracy on the OR-Bench category-matching metric. This is a calibration problem we plan to address by including OR-Bench-style benign-sensitive contrast examples in the next training cycle.

\subsection{Latency and Throughput}
\label{sec:latency}

Latency benchmarking reports throughput in samples per second and p50/p95 latency in milliseconds. Table~\ref{tab:latency_summary} summarizes the 1024-token rows, while Appendix~\ref{app:latency} reports the full latency matrix over sequence lengths 64, 256, 512, and 1024. Higher throughput and lower latency are better.

\begin{table}[H]
\centering
\small
\caption{1024-token latency and throughput summary. Higher throughput and lower latency are better.}
\label{tab:latency_summary}
\begin{tabular}{llrrr}
\toprule
\textbf{Model} & \textbf{Backend} & \textbf{Samples/s} & \textbf{p50 ms} & \textbf{p95 ms} \\
\midrule
Opir-multitask-large                  & gliclass & 50.51  & 25.65  & 26.09 \\
Opir-multitask-multilang                      & gliclass & 123.67 & 13.30  & 14.03 \\
Opir-edge                             & gliclass & \textbf{499.49} & \textbf{9.25}   & \textbf{9.52}  \\
Opir-edge-multilang                           & gliclass & 306.81 & 15.60  & 15.69 \\
GLiGuard-LLMGuardrails-300M           & gliner2  & 42.98  & 28.99  & 30.09 \\
Gliner-Guard-Omni                     & gliner2  & 34.49  & 34.04  & 34.58 \\
Llama-3.1-Nemotron-Safety-Guard-8B v3 & vllm     & 62.19  & 97.63  & 98.31 \\
PolyGuard-Qwen                        & vllm     & 23.51  & 308.59 & 309.86 \\
PolyGuard-Qwen-Smol                   & vllm     & 81.48  & 71.77  & 73.46 \\
Qwen3Guard-Gen-8B                     & vllm     & 65.45  & 91.30  & 91.80 \\
WildGuard                             & vllm     & 28.79  & 243.00 & 243.86 \\
\bottomrule
\end{tabular}
\end{table}

At 1024 tokens, \texttt{Opir-multitask-large} reaches 50.51 samples/s with 25.65/26.09~ms p50/p95 latency per sample, compared with 42.98 samples/s and 28.99/30.09~ms for GLiGuard-LLMGuardrails-300M. The binary encoder checkpoints provide the lowest latency in this run: \texttt{Opir-edge} reaches 499.49 samples/s with 9.25/9.52~ms p50/p95 latency, and \texttt{Opir-edge-multilang} reaches 306.81 samples/s with 15.60/15.69~ms p50/p95 latency. All four Opir variants are at least an order of magnitude faster at the p50 than the strongest decoder-based guardrail model in our table (Nemotron Safety Guard~v3 at 97.63~ms p50), and roughly 12--33$\times$ faster than PolyGuard-Qwen and WildGuard. This is the central practical argument for the encoder-based Opir line: comparable or better safety accuracy at one-tenth to one-thirtieth the latency of 7B--8B decoder guardrails.

\begin{figure}[H]
\centering
\includegraphics[width=0.90\textwidth]{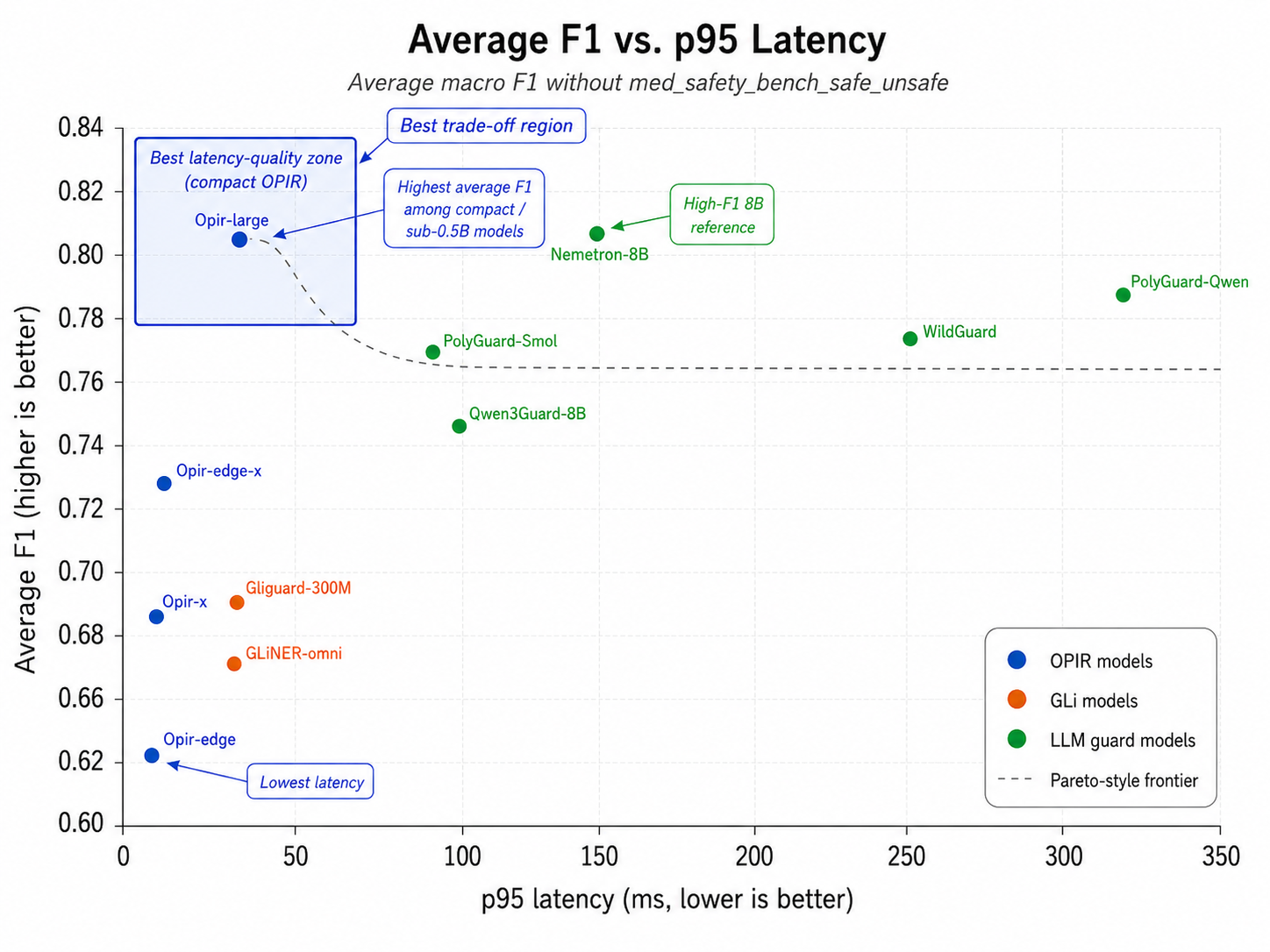}
\caption{Latency--macro-F1 efficiency comparison. The figure summarizes the trade-off between classification quality and serving cost across Opir variants and baseline guardrail systems.}
\label{fig:efficiency}
\end{figure}

\section{Limitations and Responsible Use}
\label{sec:limits}

Opir inherits several limitations common to safety classifiers. Safety labels are policy-dependent and can be subjective, especially for benign sensitive contexts, counterspeech, quoted harmful content, and educational discussion of dangerous topics; this is the open problem of \emph{over-refusal}~\citep{rottger2024xstest,cui2024orbench}. The data construction relies partly on generation and LLM-as-a-judge validation, which can introduce generator and judge biases~\citep{zheng2023judging,verga2024replacing}. The multilingual dataset is produced through translation, so performance may vary by language, dialect, code-switching pattern, and cultural context~\citep{deng2024multilingual,dewynter2024rtp}.

Additionally, it's hard to evaluate real-world, up-to-date performance, as policy standards and attack strategies can change over time. 

Opir is intended for LLM prompt and response moderation, safety routing, review prioritization, and offline safety analysis. It should not be used as the sole basis for legal, employment, credit, housing, education, law-enforcement, or other high-impact decisions, nor as a substitute for policy design, logging, appeals, human review, and abuse monitoring.

\section{Conclusion}
\label{sec:conclusion}

We presented Opir, a GLiClass-based family of encoder guardrail models for binary safe/unsafe classification, toxicity classification, jailbreak classification, and unsafe prompt or response categorization. The model family combines multi-task and edge-oriented variants with a broad three-level safety taxonomy covering 996 labels. The data construction combines taxonomy-guided synthetic generation, adversarial hard-negative mining, benign safety-preserving contrast examples, generated responses, multilingual translation, and selected public benchmark training subsets. Across an expanded comparison spanning 12 safety datasets, 17 categorization splits, and 11 contemporary guardrail systems, \texttt{Opir-multitask-large} achieves the highest average macro F1 on safety classification and wins 11 of 17 rows on categorization accuracy; the binary encoder variants reach sub-10~ms p50 latency at 1024 tokens, more than 10$\times$ faster than the strongest decoder-based baselines in our evaluations. Future work includes calibration on over-refusal benchmarks (especially OR-Bench), continual updates of the taxonomy to reflect emerging agentic threats, broader multilingual coverage, and integration of reasoning-mode classification along the lines of recent Nemotron-Content-Safety-Reasoning models~\citep{nemotronreasoning2025}.

\bibliography{main}

\clearpage
\appendix

\section{Taxonomy Detail}
\label{app:taxonomy}

This appendix lists the Level~2 subcategories and representative Level~3 leaf labels under each Level~1 category. Counts in the ``Leaves'' column are the number of Level~3 labels in the subcategory; the ``Representative leaves'' column shows a non-exhaustive sample.

\subsection*{toxicity}
\begin{small}
\begin{tabularx}{\textwidth}{lrX}
\toprule
\textbf{Subcategory} & \textbf{Leaves} & \textbf{Representative leaves} \\
\midrule
harassment\_and\_abuse & 7 & personal\_insults, targeted\_harassment, bullying, humiliation, demeaning\_language, directed\_profanity, sustained\_abuse \\
hate\_and\_discrimination & 7 & protected\_class\_slurs, identity\_based\_abuse, dehumanization, discriminatory\_stereotypes, exclusionary\_hate, supremacist\_claims, identity\_based\_threats \\
threats\_and\_intimidation & 7 & direct\_threats, implied\_threats, coercive\_intimidation, blackmail\_threats, wish\_of\_harm, menacing\_language, retaliation\_threats \\
graphic\_or\_shocking\_content & 6 & graphic\_injury\_description, gore, corpse\_desecration, celebration\_of\_suffering, animal\_cruelty\_description, shock\_content \\
abusive\_disruption & 6 & trolling, flamebait, brigading\_call, spam\_abuse, nuisance\_content, hostile\_derailment \\
psychological\_abuse\_and\_emotional\_harm & 8 & gaslighting, coercive\_control, emotional\_blackmail, isolation\_or\_dependency\_abuse, degradation\_or\_shaming, threats\_of\_abandonment, manipulative\_guilt\_or\_fear, psychological\_intimidation \\
\bottomrule
\end{tabularx}
\end{small}

\subsection*{violence\_and\_physical\_harm}
\begin{small}
\begin{tabularx}{\textwidth}{lrX}
\toprule
\textbf{Subcategory} & \textbf{Leaves} & \textbf{Representative leaves} \\
\midrule
violent\_instructions & 6 & assault\_methods, murder\_planning, torture\_methods, kidnapping\_or\_restraint, ambush\_planning, evading\_detection\_after\_harm \\
weapons\_and\_explosives & 6 & firearm\_acquisition, weapon\_modification, improvised\_weapons, explosive\_device\_construction, ammunition\_or\_ballistics, weapon\_concealment \\
public\_safety\_threats & 6 & mass\_violence\_threat, school\_or\_workplace\_threat, bomb\_threat, swatting, infrastructure\_attack, crowd\_panic\_incitation \\
extremist\_violence & 6 & terrorist\_praise, terrorist\_recruitment, attack\_planning, propaganda\_distribution, martyrdom\_encouragement, violent\_radicalization \\
dangerous\_acts & 6 & dangerous\_stunts, unsafe\_vehicle\_operation, unsafe\_workplace\_practices, tampering\_with\_safety\_equipment, encouraging\_physical\_risk, booby\_trap\_instructions \\
\bottomrule
\end{tabularx}
\end{small}

\subsection*{self\_harm\_and\_suicide}
\begin{small}
\begin{tabularx}{\textwidth}{lrX}
\toprule
\textbf{Subcategory} & \textbf{Leaves} & \textbf{Representative leaves} \\
\midrule
suicide\_risk & 6 & suicidal\_ideation, suicide\_plan, lethal\_means\_request, suicide\_encouragement, post\_attempt\_context, farewell\_or\_final\_message \\
self\_injury & 6 & cutting, burning, self\_poisoning, self\_punishment, concealing\_self\_injury, self\_harm\_challenge \\
eating\_disorders & 6 & extreme\_restriction, purging, laxative\_abuse, thinspiration, binge\_purge\_instruction, concealing\_disordered\_eating \\
acute\_distress & 6 & hopelessness, panic\_or\_crisis, trauma\_disclosure, abuse\_crisis, substance\_related\_crisis, imminent\_safety\_concern \\
harmful\_wellness\_or\_body\_practices & 6 & dangerous\_detox, unsafe\_fasting, sleep\_deprivation, extreme\_exercise, unsafe\_body\_modification, pseudomedical\_self\_treatment \\
\bottomrule
\end{tabularx}
\end{small}

\subsection*{Other categories}

For reasons of space, the remaining Level~1 categories
(\nolinkurl{sexual_content},
\nolinkurl{child_safety},
\nolinkurl{personal_information_privacy_and_intellectual_property},
\nolinkurl{cybersecurity},
\nolinkurl{criminal_and_illegal_activity},
\nolinkurl{regulated_goods_and_advice},
\nolinkurl{biological_medical_and_environmental_harm},
\nolinkurl{weapons_of_mass_destruction},
\nolinkurl{information_integrity_and_manipulation},
\nolinkurl{ai_system_security_and_reliability},
\nolinkurl{bias_fairness_and_representation},
\nolinkurl{other_or_uncertain}, and
\nolinkurl{safe_and_benign})
follow the same Level~2 / Level~3 layout.

The complete taxonomy, including all 126 Level~2 subcategories
and all 854 Level~3 leaf labels, is shipped with the released models. 

Notable design choices include:
an 18-subcategory
PII/IP
branch with 129 leaves covering both PII exposure and
surveillance/drone misuse;
a 22-subcategory
medical and environmental
branch with 177 leaves covering pathogen access,
gain-of-function research, lab-safety failures,
and dual-use research escalation; and an explicit
AI system and security
branch covering instruction-hierarchy attacks~\citep{wallace2024instruction},
indirect prompt injection~\citep{greshake2023not},
tool/connector abuse, and unsafe autonomy.

\section{Latency and Throughput Details}
\label{app:latency}

Table~\ref{tab:latency_full} reports the full latency and throughput matrix from the benchmark log. The log includes GLiClass, GLiNER2, and vLLM guardrail backends across sequence lengths 64, 256, 512, and 1024. Hardware and serving-configuration metadata are not present in the log, so these values should be interpreted as within-run measurements.

\begin{table}[p]
\centering
\footnotesize
\caption{Full latency and throughput matrix across sequence lengths 64, 256, 512, and 1024.}
\label{tab:latency_full}
\begin{tabular}{llrrrr}
\toprule
\textbf{Model} & \textbf{Backend} & \textbf{Seq.} & \textbf{Samples/s} & \textbf{p50 ms} & \textbf{p95 ms} \\
\midrule
best (Opir-multitask-large) & gliclass & 64   & 354.22  & 21.13 & 21.25 \\
best                         & gliclass & 256  & 329.78  & 21.64 & 21.74 \\
best                         & gliclass & 512  & 139.40  & 22.43 & 22.82 \\
best                         & gliclass & 1024 & 50.51   & 25.65 & 26.09 \\
\midrule
multi\_best (Opir-multitask-multilang) & gliclass & 64   & 646.06  & 10.85 & 10.96 \\
multi\_best                    & gliclass & 256  & 582.32  & 11.98 & 12.19 \\
multi\_best                    & gliclass & 512  & 341.31  & 12.71 & 13.44 \\
multi\_best                    & gliclass & 1024 & 123.67  & 13.30 & 14.03 \\
\midrule
bi\_en\_best (Opir-edge)       & gliclass & 64   & 1024.53 & 6.49  & 6.54  \\
bi\_en\_best                   & gliclass & 256  & 836.41  & 7.38  & 7.74  \\
bi\_en\_best                   & gliclass & 512  & 740.11  & 7.80  & 7.88  \\
bi\_en\_best                   & gliclass & 1024 & 499.49  & 9.25  & 9.52  \\
\midrule
bi\_multi\_best (Opir-edge-multilang)  & gliclass & 64   & 556.95  & 12.96 & 13.06 \\
bi\_multi\_best                & gliclass & 256  & 525.04  & 13.61 & 13.68 \\
bi\_multi\_best                & gliclass & 512  & 471.71  & 14.33 & 14.43 \\
bi\_multi\_best                & gliclass & 1024 & 306.81  & 15.60 & 15.69 \\
\midrule
gliguard-LLMGuardrails-300M    & gliner2 & 64   & 449.70 & 11.24 & 11.99 \\
gliguard-LLMGuardrails-300M    & gliner2 & 256  & 182.15 & 13.48 & 14.63 \\
gliguard-LLMGuardrails-300M    & gliner2 & 512  & 90.84  & 16.03 & 16.70 \\
gliguard-LLMGuardrails-300M    & gliner2 & 1024 & 42.98  & 28.99 & 30.09 \\
\midrule
gliner-guard-omni              & gliner2 & 64   & 412.69 & 11.12 & 11.19 \\
gliner-guard-omni              & gliner2 & 256  & 160.91 & 13.35 & 13.48 \\
gliner-guard-omni              & gliner2 & 512  & 78.51  & 17.13 & 17.72 \\
gliner-guard-omni              & gliner2 & 1024 & 34.49  & 34.04 & 34.58 \\
\midrule
Llama-3.1-Nemotron-SG-8B-v3    & vllm    & 64   & 71.70  & 94.77  & 95.83 \\
Llama-3.1-Nemotron-SG-8B-v3    & vllm    & 256  & 70.73  & 95.29  & 95.77 \\
Llama-3.1-Nemotron-SG-8B-v3    & vllm    & 512  & 64.24  & 95.86  & 96.09 \\
Llama-3.1-Nemotron-SG-8B-v3    & vllm    & 1024 & 62.19  & 97.63  & 98.31 \\
\midrule
PolyGuard-Qwen                 & vllm    & 64   & 24.20  & 309.39 & 314.54 \\
PolyGuard-Qwen                 & vllm    & 256  & 24.08  & 305.14 & 311.62 \\
PolyGuard-Qwen                 & vllm    & 512  & 24.31  & 306.51 & 307.09 \\
PolyGuard-Qwen                 & vllm    & 1024 & 23.51  & 308.59 & 309.86 \\
\midrule
Qwen3Guard-Gen-8B              & vllm    & 64   & 75.59  & 88.52  & 89.72 \\
Qwen3Guard-Gen-8B              & vllm    & 256  & 75.28  & 89.14  & 90.89 \\
Qwen3Guard-Gen-8B              & vllm    & 512  & 73.97  & 89.66  & 91.76 \\
Qwen3Guard-Gen-8B              & vllm    & 1024 & 65.45  & 91.30  & 91.80 \\
\midrule
PolyGuard-Qwen-Smol            & vllm    & 64   & 96.32  & 69.84  & 72.27 \\
PolyGuard-Qwen-Smol            & vllm    & 256  & 93.22  & 71.92  & 73.37 \\
PolyGuard-Qwen-Smol            & vllm    & 512  & 90.40  & 70.80  & 72.00 \\
PolyGuard-Qwen-Smol            & vllm    & 1024 & 81.48  & 71.77  & 73.46 \\
\midrule
wildguard                      & vllm    & 64   & 31.68  & 242.55 & 245.96 \\
wildguard                      & vllm    & 256  & 30.74  & 239.16 & 239.66 \\
wildguard                      & vllm    & 512  & 30.49  & 240.13 & 241.46 \\
wildguard                      & vllm    & 1024 & 28.79  & 243.00 & 243.86 \\
\bottomrule
\end{tabular}
\end{table}

\clearpage

\end{document}